# Metacognitive Learning Approach for Online Tool Condition Monitoring

Mahardhika Pratama[a], Eric Dimla[b], Chow Yin Lai[c], Edwin Lughofer[d]

[a] School of Computer Science and Engineering, Nanyang Technological University, Singapore, pratama@ieee.org
[b] Mechanical Engineering Programme Area, Faculty of Engineering, Universiti Teknologi Brunei, Jalan Tungku Link, Gadong BE1410, Bandar Seri Begawan, Brunei Darussalam, dimla@utb.edu.bn
[c] School of Engineering, RMIT University, Carlton, 3053, Victoria, Australia, chowyin.lai@rmit.edu.au
[d] Department of Knowledge-Based Mathematical Systems, Johannes Kepler University, Linz, Austria, edwin.lughofer@jku.at

**Abstract**
As manufacturing processes become increasingly automated, so should tool condition monitoring (TCM) as it is impractical to have human workers monitor the state of the tools continuously. Tool condition is crucial to ensure the good quality of products – Worn tools affect not only the surface quality but also the dimensional accuracy, which means higher reject rate of the products. Therefore, there is an urgent need to identify tool failures before it occurs on the fly. While various versions of intelligent tool condition monitoring have been proposed, most of them suffer from a cognitive nature of traditional machine learning algorithms. They focus on the how-to-learn process without paying attention to other two crucial issues – what-to-learn, and when-to-learn. The what-to-learn and the when-to-learn provide self-regulating mechanisms to select the training samples and to determine time instants to train a model. A novel tool condition monitoring approach based on a psychologically plausible concept, namely the metacognitive scaffolding theory, is proposed and built upon a recently published algorithm – recurrent classifier (rClass). The learning process consists of three phases: what-to-learn, how-to-learn, when-to-learn and makes use of a generalized recurrent network structure as a cognitive component. Experimental studies with real-world manufacturing data streams were conducted where rClass demonstrated the highest accuracy while retaining the lowest complexity over its counterparts.

*Keyword – Prognostic Health Management, Online Learning, Evolving Intelligent System, Lifelong Learning, Nonstationary Environments, Concept Drifts*

## 1. Introduction

To survive in the current competitive manufacturing industry, a company must improve its manufacturing productivity and product quality. Automation is a natural way forward, as automated machines and robots can produce parts with high precision and continuously without fatigue. Human workers can then be re-trained to perform less laborious and less dangerous tasks such as programming the machines or supervising the overall workflow.

For machining operations with cutting tools such as drilling, milling or turning, "automation" involves machine tending (loading and unloading of parts), the actual cutting operation itself, and tool condition monitoring. The latter is extremely important because worn tools affect surface quality and dimensional accuracy (Dimla et al., 1997; Rehorn et al., 2005; Ambhore et al., 2015), thereby increasing reject rates and wastage, creating a detrimental effect on both the profit and image of the company. Furthermore, the cutting force needs to be increased when the tool is blunt, which translates to higher energy cost. On the whole, in the worst-case scenario, a catastrophic failure of the machine tool would result in a significant cost incur as replacement parts or whole machines have a very high capital outlay.

While replacing worn tools too late is undesired, the other extreme of replacing tools to sharper ones too early and too frequently is also not a good idea, as this increases the tool costs and downtime related to tool replacement. As such, the ideal method is to have an online tool condition monitoring (TCM) system which constantly detects the state of the tool, and initiates a tool change only when this is required.

An automated TCM system needs to perform two main tasks, namely sensing and monitoring (Elbestawi et al., 2006). Sensing means using sensors to obtain cutting process signals, and this can further be divided into two categories – direct or indirect sensing (Ambhore et al., 2015; Elbestawi et al., 2006). Indirect sensing method, the tool or the workpiece is analyzed at the end of the machining cycle, through optical measurement, surface-finishing measurement, chip size measurement, etc. Disadvantages of this approach include production time loss for the purpose of measurement, as well as the inability to detect tool failure between measurements at an early stage before the failure becomes visible on the final product. Thus, it cannot prohibit any production waste, but just sort out bad items and waste in an a posteriori fashion. The indirect sensing technique, on the other hand, uses correlated variables such as machining force, spindle current, acoustic emission and vibration to determine the status of the tools. It is intuitively clear that as the tool becomes blunt, it takes higher force and spindle current to cut through the metal. Therefore, by sensing these signals, one can have a good estimate of the tool condition.

The second task, i.e. monitoring, involves processing the measured signals and performing classification or decision making. Signal processing may include sensor fusion (Khaleghi et al., 2013), transforming the time-domain signal into frequency domain data using Fourier Transformation (Ding, 2008), statistical analysis, etc. Lastly, the decision-making process will determine whether a tool is still usable or replacement is required. The earliest and simplest method is to trigger a tool change when certain threshold values are exceeded (Elbestawi et al., 2006). Significant research efforts have been put into this area to make the decision making more reliable, for e.g. making systems which have the ability to learn and adapt to new information, extract features, learn from their experience (e.g. false alarm) and changes themselves to be more robust. This is done using a data-driven approach where a machine learning algorithm such as NN, etc (Dimla et al., 1997) is deployed to learn a set of manufacturing data.

Neural networks are mathematical processing devices consisting of some highly interconnected elements, and they are excellent in performing nonlinear modeling, function approximation and pattern classification (Dimla et al., 1997). They are also capable of handling a large amount of data. Because of these strengths, they are a natural candidate for the decision-making task of TCM systems. A survey in 1997 (Dimla et al., 1997) reveals that there were already at least 40 publications discussing the use of neural networks for TCM, and this number continues to rise due to better computing power and newer algorithms for the training of the neural networks. Interested readers are also referred to (Elbestawi et al., 2006) for a more recent survey in 2006.

It is understood from the literature that the vast majority of data-driven TCMs works in a batch mode due to the use of an offline machine learning algorithm. This drawback hinders their viability to deal with sensory data streams collected in an online real-time fashion from in-service machinery. Moreover, data are sampled in a very fast rate in modern manufacturing processes with limited interruption only. It is inter-correlated and means any shutdown leads to a complete shutdown to the overall production cycle. Existing works in the TCM have a limited capability to cater different process parameters because they have a fixed model capacity (Pratama et al., 2013).

Recent advances in the area of Evolving Intelligent System (EIS) (Angelov et al., 2010, Mouchaweh. Lughofer, 2012) offers a promising breakthrough for the TCM because it strengthens the so-called "maintenance on-demand" paradigm (Scheffer, 2004) which makes continuous monitoring of tool condition on the fly possible, while engaging the cutting process. The salient feature of the EIS is perceived in its online learning trait providing an effective avenue in dealing with data streams and appropriately updating models in dynamically changing, non-stationary processes with low memory complexity and fast training speed. They play vital roles to the success of the TCM in the high-speed milling process. The EIS adopts an open structure, which can start its training process from scratch with an empty rule base. Its structure is self-evolved in the single-pass learning mode, which suits to the degree of nonlinearity and non-stationary characteristics of a system (Lughofer, 2011). This trait brings significant advantage to the TCM because the machining parameters vary to meet production requirements. Furthermore, cutter degradation often leads to a gradual change of machining behaviors which demand a self-organizing scenario to adapt to such conditions. For real-time deployment, the EIS, however, carries some bottlenecks as a result of its cognitive trait. The vast majority of existing EISs are crafted in the fully supervised training scenario which results in costly labeling cost. On top of that, current EISs have not characterized a plug-and-play characteristic due to the absence of important learning modules making them over-dependent on pre-and/or post-processing steps (Pratama et al, 2015(b)).

This paper presents a metacognitive learning approach for an online tool condition monitoring. A novel TCM is developed using a recently published metacognitive scaffolding learning machine, termed Recurrent Classifier (rClass) (Pratama et al., 2015(b)). rClass actualises three components of metacognition in psychology, namely what-to-learn, how-to-learn, when-to-learn, into the machine learning context with sample deletion strategy, sample learning strategy, sample reserved strategy respectively (Suresh et al., 2010). The sample deletion strategy is developed under the roof of an online active learning scenario – the conflict measure – which puts into perspective the semi-supervised working principle (Lughofer, 2013). The semi-supervised working principle comes into the picture because it prevents continuous labeling of data streams which leads to a significant reduction of execution time, annotation efforts by an operator as well as an improvement of model's generalization. Also, the original version of the what-to-learn module of rClass is enhanced here to address the class imbalanced issue. The how-to-learn scenario is constructed under the Scaffolding theory – a prominent tutoring theory for a learner to study a complex learning task (Wood, 2001). The Scaffolding theory supports the plug-and-play paradigm because of its three components: complexity reduction, fading and problematizing. The complexity reduction aims to simplify learning complexity and is realized with the input weighting scenario to cope with the curse of dimensionality. The problematizing is meant to address concept drifts in the data stream and includes rule growing scenario to cope with the sudden concept drift, rule forgetting scenario to handle the gradual concept drift, rule splitting scenario to address the incremental concept drift and rule recall scenario to overcome the cyclic concept drift. Fading scenario targets model's complexity and is implemented using the rule pruning scenario.

rClass is developed under a generalized recurrent network structure where it combines a generalized Takagi-Sugeno-Kang (TSK) fuzzy rule and a local recurrent network structure (Juang et al., 2010). The local recurrent network structure is put forward to overcome dependency on time-delayed input variable and temporal system dynamic while maintaining the local property playing a key role to flexibility and stability of EIS. The generalized TSK fuzzy system synergies the multivariate Gaussian function as the rule premise while integrating a Chebyshev polynomial up to a second order. The multivariate Gaussian function aims to generate a non-axis parallel ellipsoidal cluster while the Chebyshev function is meant to rectify the mapping capability of first order TSK polynomial (Patra, 2002). The metacognitive TCM methodology here conveys the following advantages over existing approaches: 1) it works fully in the single-pass learning mode, which renders it compatible for online real-time deployment; 2) it adopts an open structure which adapts flexibly to any variations of data streams. Furthermore, the problematizing of the Scaffolding theory helps to handle various concept drifts; 3) it incurs low labelling cost by operators and is seen as a semi-supervised machine in some sense; 4) it realises the plug-and-play working principle where all learning modules are embedded in a single training process without any pre-and/or post-training steps; 5) its recurrent structure makes possible to cope with the temporal characteristic of a machining process. It is evident that many real-world industrial processes feature strong temporal dependencies among subsequent patterns whereas a feedforward network architecture discounts the order of data presentation unless with the use of time-delayed input variables; 6) a modification of the what-to-learn part in this paper makes possible to cope with the class imbalanced issue.

The contributions of this paper are summed up as follows: 1) this paper presents a novel TCM inspired by prominent concepts of Metacognitive Scaffolding learning; 2) the what-to-learn component of rClass is generalized here to overcome the class imbalance problem; 3) Real-time experiment using a real-world manufacturing plant was done where raw manufacturing data were collected, pre-processed; 4) The efficacy of a novel TCM was numerically validated using real-world manufacturing data. It was also compared against state-of-the-art algorithms. It is shown that our algorithm outperformed its counterparts in both accuracy and complexity.

The remainder of this paper is structured as follows: Section 2 outlines literature survey over existing TCMs, Section 3 elaborates learning policy of rClass, Section 4 discusses problem description of a real-world manufacturing plant, Section 5 describes our experiments and comparisons with prominent algorithms, concluding remarks are drawn in the last section of this paper.

## 2. Literature Survey
*A. Tool Wear*

Cutting tool wear can be classified into several types as follows (Dimla, 2000):

i. Adhesive wear associated with shear plane deformation,
ii. Abrasive wear resulting from hard particles cutting action,
iii. Diffusion wear occurring at high temperatures, and
iv. Fracture wear such as chipping due to fatigue.

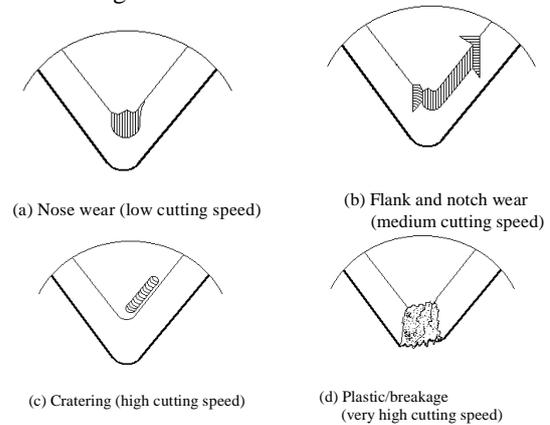

(a) Nose wear (low cutting speed)   (b) Flank and notch wear (medium cutting speed)
(c) Cratering (high cutting speed)   (d) Plastic/breakage (very high cutting speed)

Figure 1: Cutting tool wear forms in orthogonal metal cutting (Dimla, 2000)

Tool wear processes occur in combination with the predominant wear mode, dependent upon the cutting conditions, workpiece and tooling material, and the tool inserts geometry. For a given cutting tool and workpiece material combination, the tool wear form may depend exclusively on the cutting conditions, principally cutting speed $V$ and the undeformed chip thickness $t$, and a combination of the wear mechanisms above. Ranges of cutting speed where each type of wear is predominant can be identified by considering the product of these values as $Vt$ which is proportional to the cutting speed (Shaw, 1984). Sometimes, the tool life can be considerably reduced if the area of cut, the area swept by the cutting tool, is significantly increased (i.e. by increasing the depth of cutting mainly). At low cutting speeds, the tool wears predominantly by a rounding-off of the cutting point and subsequently loses sharpness. As the cutting speed increases the wear-land pattern changes to accommodate the ensuing change with extremely high values leading to plastic flow at the tool point. The various forms of wear-land pattern and prevailing cutting speed in Figure 1. The more predominantly occurring forms of cutting tool wear often identified as the principal types of tool wear in metal turning using single-point tools are the nose, flank, notch and crater wear. Figure 2 shows how wear features are measured.

Nose Wear or edge rounding occurs predominantly through the abrasion wear mechanism on the cutting tool's major edges resulting in an increase in negative rake angle. Nose wear can be dependent entirely on the implemented cutting conditions with tool sharpness lost through plastic or elastic deformation. At high cutting speeds, the edge deforms plastically and may result in the loss of the entire nose, Figure 1(a) and Figure 2(b). Edge chipping and cracking occur during periodic breaks of the 'built up edge' in interrupted cuts with the brittle tool and thermal fatigue.  Catastrophic failure may also occur if the nose is considerably worn or as a result of the utilization of inappropriate machining conditions and brittle tools such as ceramics and cemented carbide (Schey, 1987).

Flank wear arises due to both adhesive and abrasive wear mechanisms from the intense rubbing action of the two surfaces in contact, i.e. the clearance face of the cutting tool and the newly formed surface of the workpiece.  Its rate of increase at the beginning of the tool life is rapid, settling down to a steady-state then accelerating rapidly again at the end of tool life (Figure 3). Note that this figure just presents a rough estimated of the tool wear time cycle during the machining process. The tool wear does not occur in a specific interval time.  Flank wear leads to a deterioration of surface quality, increased contact area and consequently to increased heat generation (Figure 1(b) and Figure 2(c)). Wear notch forms at a depth of cut line as the tool rubs against the shoulder of the workpiece (Figures 1(b) and 2(c)). Wear notch can lead to abrasion setting by the surface layers accelerated by oxidation or chemical reactions, possibly leading to total tool failure.

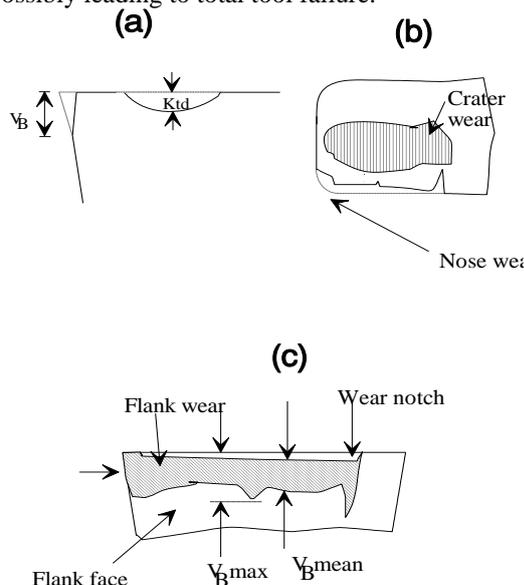

Figure 2:  Conventional features of tool wear measurements (Dimla, 2000)

Crater wear results from a combination of high cutting temperatures and high shear stresses creating a crater on the rake face some distance away from the tool edges, quantified by the depth and cross-sectional area (Fig. 1(c)). Crater wear also arises due to a combination of wear mechanisms: adhesion, abrasion, diffusion or thermal softening, and plastic deformation. Severe depths of Crater may trigger a catastrophic collapse of the cutting point (Fig. 1(d)).

*B. Tool Condition Monitoring*

Techniques for on-line TCM system can be grouped into two main categories: ***direct sensing*** and ***indirect sensing*** techniques.  While direct methods of wear measurement have been attempted, the majority of methods have been indirect (Dimla 2000; Rangwala and Dornfeld, 1990;).

Direct methods are of less benefit because the cutting area is largely inaccessible, and therefore monitoring cannot be performed while the tool is actively engaged in in-process cutting.  These methods include amongst others, touch trigger probes, optical, radioactive, proximity sensors and electrical resistance measurement techniques.

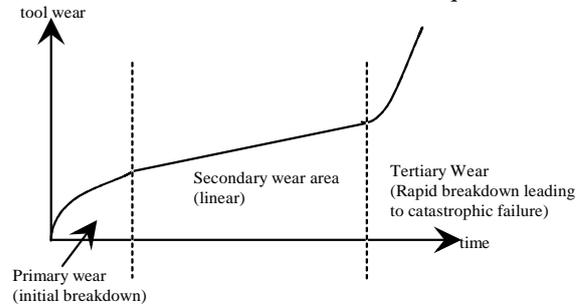

Figure 3:  Generalized tool flank and nose wear progression (Dimla, 1998)

Indirect methods take measurements while the tool is actively engaged since it involves recording a variable that can be correlated to tool wear (i.e. indirect methods measure factors that result as a consequence of tool wear).  Commonly used methods include amongst others cutting forces, acoustic emission, temperature, vibration, spindle motor current, cutting conditions, torque, and strain. These factors reflect far more than tool wear alone and parameters associated with tool wear must, therefore, be extracted from them and correlated to give a measure or extent of tool wear.  The main practical drawback with this popular method is the need for calibration of the associated parameters in monitoring the cutting process.  The cutting conditions (speed, feed-rate, and depth of cut) are known to affect the sensor signals and a range of methods have been suggested for separating the effects of these conditions from those of wear on the measured parameter (Karmathi, 1994).

Methods for correlating the measured process parameters to tool wear, breakage or chipping fall primarily into three categories.  The first class consists of methods that can be viewed as ***heuristic based rules*** with a priori knowledge only of the process parameters, and use as mathematical modeling and adaptive observers.  The second category consists of methods that could be viewed as requiring formal knowledge of the process, and these can be grouped together as ***analytically based models*** such as time series analyses and Fast Fourier Transform peak tracking.  The last category is one of example based models with ***inductive learning capabilities*** such as pattern recognition, decision surfaces, mapping techniques, clustering, and ANN.

**Methods for Detecting Tool Wear**

Most indirectly measured sensor signals are affected by workpiece material variation, the geometry of the cutting tool and the cutting conditions (Lister, 1993). An on-line monitoring system designed to take the above factors into consideration is extremely demanding. In many cases, the machining system does not need to have a global knowledge of the machining process, as it only requires knowledge of the machining operation in the neighborhood of the optimum for the operating process (Byrne *et al.*, 1995). The cutting conditions cannot be neglected, as their variable parameters affect the sensor signals and therefore the cutting tool state.

The use of these indirect methods requires accurate predictive models that link the un-measurable parameter of interest (tool wear), with the indirectly detectable variables (Oraby and Hayhurst, 1991). The process of tapping useful signals or process parameters directly or indirectly from a cutting process constitutes what Burke (1989) symbolically terms, the ***detection level***. This level in any TCMS is primarily concerned with choosing the relevant sensory signal.

**Identification of Tool Wear**

The choice of a suitable sensor type and its point of application (sensor location) are inextricably linked, with a suitable location for any such sensor being where the specified signal has the highest concentration and best reproducibility. Certain limitations, however, do exist due to the structure of the machine. A school of thought (Dimla, 1998) has it that information from a variety of different sensors has to be collected and these signals of varying reliability integrated (fused). The primary benefit of sensor fusion would seem the fact that it serves to enhance the richness of the underlying information (wear level) contained in each signal thereby increasing the reliability of the monitoring process (Dornfeld, 1990). The lack of an accurate model for wear prediction has thus led researchers resorting to other methods of sensor integration which do not rely on any theoretical or empirical process.

The fusion and subsequent correlation of sensor signal features to tool wear is viewed as constituting the identification of wear, thus the characteristic ***identification level*** (Burke, 1989). The utilization of single sensor signals offers incomplete sensory information as to the wear state (condition) of an engaged cutting tool. This shortfall can be obviated by pursuance of a sensor fusion strategy (Lui and Ko, 1990). This method has proven highly successful as more reliable information on the tool state was extractable. In addition to pattern recognition techniques, hyperplane decision surfaces mapping techniques, neural networks, clustering techniques and perceptron techniques are reported by Noori-Khajavi and Komanduri (1995) in their literature survey as other fusion techniques. Methods primarily based on neural networks have proven particularly popular because tool wear intrinsically is a non-linear process. Its representation via a neural network is viable, reliable and an attractive alternative to previously employed empirical methods (Elanayar *et al.*, 1990). The proposal of EISs - a successor of standard neural network for online and dynamic environments - has attracted growing research interest in the TCM field. Li et al, (2010) come up with the fuzzy regression technique for tool performance prediction and degradation detection. Wang et al, (2010) applied the so-called FAOS-PFNN for tool wear prediction of ball nose end-milling cutter. Nonetheless, they work in the offline mode which imposes a prohibitive computational cost for the strictly online environment. A fully online approach for predictive analytics of tool wear was proposed in (Oentaryo et al, 2011). The so-called DFNN was deployed for machine health condition prediction by Pan et al, (2014). To date, existing TCM systems rely on conventional EIS variants. Such systems are cognitive nature since two important facets, namely what-to-learn and when-to-learn, are uncharted.

**3. Evolving Intelligent Systems – Theory and Algorithm**

*A. Cognitive Component of rClass*

Recurrent Classifier (rClass) is driven by a generalized Takagi-Sugeno-Kang (TSK) fuzzy system where the premise part is built upon a multivariable Gaussian function (Pratama et al., 2014(a)), while the rule consequent is constructed with the Chebyshev polynomial up to the second order (Patra et al., 2002). This fuzzy rule is structured under a local recurrent network structure generating the spatiotemporal trait.

The advantage of the Chebyshev polynomial over the standard zero- or first-order TSK rule consequent lies in a better approximation of a local output behavior of a nonlinear functions (Patra et al., 2002) because it features a higher Degree of Freedom. It is also more convenient to compute than the trigonometric function (Pao, 1989) because it involves less number of parameters as follows:

$$T_{n+1}(x_j) = 2x_j T_n(x_j) - T_{n-1}(x_j) \qquad (1)$$

If the second order is the order of interest and we deal with a 2-D input space $[x_1, x_2]$. The extended input vector $x_e$ is formed as follows:

$$x_e = [1, T_1(x_1), T_2(x_1), T_1(x_2), T_2(x_2)] \qquad (2)$$

where $x_e \in \Re^{1 \times (2u+1)}$ and $u$ is the number of input dimension. It is worth noting that the zero or first-order TSK rule consequent suffers from a high bias problem which does not suffice in modelling a highly nonlinear characteristic.

The rule premise of the generalized TSK fuzzy rule relies on the multivariate Gaussian function with a non-diagonal covariance matrix (Pratama et al., 2014(a)). This function generates a non-axis parallel ellipsoidal cluster which rotates to any direction. This trait lowers fuzzy rule demand which compensates a possible increase of network parameters as a result of the non-diagonal covariance matrix. The multivariate Gaussian function is expressed as follows:

$$R_i = \exp(-(\lambda X_n - C_i)\Sigma_i^{-1}(\lambda X_n - C_i)^T) \qquad (3)$$

where $X_n \in \mathcal{R}^{1\times u}$ and $\lambda \in \mathcal{R}^{1\times u}$ are respectively an input vector of interest and an input weight vector. The input weight vector is obtained from the online input weighting strategy based on the L1 LDA method which is meant to induce the soft-dimensionality reduction approach. $C_i \in \mathcal{R}^{1\times u}$ is a Centroid of the *i-th* rule and $\Sigma_i^{-1} \in \mathcal{R}^{u\times u}$ is an inverse non-diagonal covariance matrix whose elements govern orientation of a non-axis parallel ellipsoidal cluster. Equation (3) produces a spatial firing strength of a fuzzy rule which also defines a compatibility degree of a data point but excludes the temporal characteristic of the problem. This downside is overcome by putting forward a local recurrent network structure generating a spatio-temporal firing strength as follows:

$$\varphi_{i,o} = \gamma_{i,o} R_i + (1 - \gamma_{i,o})\varphi_{i,o}(n-1) \quad (4)$$

where $\gamma_{i,o}$ is a local recurrent weight of the *i-th* rule and *o-th* class and also controls a tradeoff between previous and current information. The rClass's output is defined as a weighted average of the spatio-temporal firing strength and the rule consequent:

$$y_o = \frac{\sum_{i=1}^{M}\varphi_{i,o}\Psi_{i,o}}{\sum_{i=1}^{M}\varphi_{i,o}} = \frac{\sum_{i=1}^{M}\varphi_{i,o} x_e W_{i,o}}{\sum_{i=1}^{M}\varphi_{i,o}} \quad (5)$$

where $M$ is the number of fuzzy rules and $W_{i,o} \in \mathfrak{R}^{(2u+1)\times 1}$ is the output weight vector of the *i-th* rule and *o-th* class. The normalization in equation (5) is to assure the partition of unity. The classification decision is inferred using the so-called MIMO architecture as follows:

$$O = \max_{o=1,\dots,C}(y_o) \quad (6)$$

where $C$ is the number of classes. The MIMO architecture is deemed reliable in predicting a class label in an overlapping region because each rule features different rule consequents per class.

*B. Metacognitive Component of rClass*

This section outlines the metacognitive part of rClass which consists of three components: what-to-learn, how-to-learn and when-to-learn.

*B.1 What-To-Learn*

Unlike the sample deletion strategy (Suresh et al., 2010), the online active learning strategy is incorporated in rClass and contributes to significant relieve of the operator's annotation efforts because sample contribution is estimated with the absence of labeled samples. It brings rClass's working principles to go one step ahead to the semi-supervised principle (Xiong et al., 2014).

rClass makes use of the so-called Bayesian conflict method to run the what-to-learn strategy. This method offers more advanced traits than some of its counterparts in (Lughofer et al., 2013; Pratama et al., 2014(c)) because it features a dynamic conflict level which adapts to speed and severity of concept drifts (Zlibaite, 2014). It takes into account the overlapping class problem because it utilizes the class-prior probability in measuring the conflict level. The conflict is formalised in both input and output spaces to cope with the so-called virtual and real concept drifts. The conflict in the output space is measured by the Bayesian posterior probability:

$$p(y_{o,output}|X) = \min(\max(conf,0),1) \quad (7)$$
$$conf = \frac{y_1}{y_1+y_2} \quad (8)$$

where $y_1, y_2$ are the most and second most dominant output classes of rClass. It is observed from (7), (9) that the conflict in the output space is set as a spatial proximity between a target sample to a decision boundary. A sample sitting adjacent the decision boundary incurs a high conflict which pinpoints the need for a reshape of decision boundary. The conflict in the input space is measured using the Bayesian posterior probability as follows:

$$P(y_{o,input}|X) = \frac{\sum_{i=1}^{M} P(y_o|R_i)P(X|R_i)P(R_i)}{\sum_{o=1}^{C}\sum_{i=1}^{M} P(y_o|R_i)P(X|R_i)P(R_i)} \quad (9)$$

where $P(y_o|R_i), P(X|R_i), P(R_i)$ are the class prior probability, the likelihood function and the prior probability. The advantage of (9) over a standard compatibility measure as shown in (Lughofer, 2013) is seen in its class prior probability which enables unclean clusters to be handled properly. That is, the unclean cluster portrays a situation where a cluster is occupied by different-classes samples. Unclean samples are inherent to misclassifications because one cluster cannot be assigned with a unique class. The class posterior probability, the prior probability and the likelihood function (Vigdor et al, 2007) are written as follows:

$$P(y_o|R_i) = \frac{\log(N_{i,o}+1)}{\sum_{o=1}^{C}\log(N_{i,o}+1)} \quad (10)$$

$$P(R_i) = \frac{\log(N_i+1)}{\sum_{i=1}^{M}\log(N_i+1)} \quad (11)$$

$$P(X|R_i) = \frac{1}{(2\pi V_i)^{1/2}}\exp(-(X-C_i)\Sigma_i^{-1}(X-C_i)^T) \quad (12)$$

The log operator is added to the class prior probability and the prior probability to soften its effect to a new rule. A sample is admitted for the training process provided its conflicts in both input and output spaces exceed a conflict threshold $\theta$:
$$P(y_{o,input}|X) and P(y_{o,output}|X) < \theta \quad (13)$$

To avoid having expensive labeling cost, a budget $B$ is set and determines the maximum number of annotations during the training process. The training process is terminated, and a model is fixed if the actual labeling effort calculated as $b = Z/N$ goes beyond this level where $Z$ is the number of annotations made so far and $N$ is the number of queried samples. This strategy plays important roles in real-world data stream cases because the labelling effort can be intractable due to infinite nature of data streams. The actual budget $b$ can be insensitive when the number of samples is large. A slight modification is made for the actual labelling cost as follows:

$$Z(n+1) = Z(n)o + labelling$$
$$o = \frac{(\xi-1)}{\xi}, b = \frac{Z}{\xi} \quad (14)$$

where $\xi$ is the window size, fixed at 100 in all our simulations in this paper and *labelling* signifies whether a sample of interest is labelled or not. In addition to (13), another criterion to accept a data sample is added where a sample is labelled only when a budget is not fully spent as follows:
$$b \leq B \quad (15)$$
Because of non-stationary learning environments, the conflict threshold should not be kept constant. Otherwise, it results in running out the budget too quickly since concept drifts impose a high labeling demand due to uncertainties in data distributions. Assuming that $h = P(y_o|X)$ is a random variable with the minimum $1/C$ and the maximum 1, we can derive a good estimate of the conflict threshold under the constraint $b \leq B$ as follows:
$$P(h \leq \theta) = P\left(h \in \left[\frac{1}{C}, 1\right]\right) = \int_{1/C}^{1} p(h)dh \leq B$$
The final formula is reliant on the assumption of data distributions. Given that the uniform distribution is applied, we arrive at the following expression.
$$\theta = \frac{1}{C} + B\left(1 - \frac{1}{C}\right) \quad (16)$$
This formula is further adjusted to adapt to concept changes in data streams. The adaptive strategy is also required to compensate the labeling cost in the presence of concept drifts and is formalized as $\theta = \theta(1 \pm s)$ where $s$ is assigned its default value $s=0.05$ (Zliobaite et al, 2..014). The conflict threshold increases $\theta(1 + s)$ when a sample is discarded – (13) and (16) are not satisfied, whereas it diminishes $\theta(1 - s)$ when a data point is accepted for the training process – (13) and (16) are satisfied.

The so-called class imbalance problem has to be addressed in the online active learning scenario because under this circumstance a classifier has poor generalization performance for an under-represented class although it delivers a decent overall classification rate. The active learning scenario of rClass is generalized here to take into account the class imbalance problem. A minority-class sample is given a priority to be accepted for model updates and this is done by inspecting the estimated class-posterior probability in the input and output space:
$$\max_{o=1,..,C} P(y_{o,in}|X) = \max_{o=1,...,C} P(y_{o,out}|X) = \hat{o} \quad (17)$$
where $\hat{o}$ is a minority class. This situation implies a queried sample most likely lies in a minority target class. Such samples should not be ignored to prevent the decision boundary to be skewed toward a majority class. This scenario is activated given that the class imbalance is detected. That is, the imbalance factor is defined as follows:
$$IF = 1 - \frac{C}{N} \min_{o=1,..,C} N_o \quad (18)$$
where the higher the value reveals the more severe the class imbalance exists (Subramanian et al, 2014(a)). The imbalance factor is set at $IF \geq 0.3$ to turn on (17) which shows about 15% of total samples belong to a minority class. The minority class is set as that of $\hat{o} < 0.3N$.

*B. 2 How-to-Learn*

This section focuses on the how-to-learn part which updates the cognitive component and is derived from the Scaffolding theory (Wood et al., 2001).

• *Rule Growing Strategy*: rClass is equipped by two rule growing scenarios, namely Datum Significance (DS) and Data Quality (DQ) methods (Pratama et al., 2015(b)) which coexist to estimate the significance of training samples. The DS method estimates the statistical contribution of training samples which reflects their possible contribution to the overall training process. The DS method is devised under the uniform data distribution assumption, which ignores the zone of influence of training samples in the feature space. The DQ method is incorporated to overcome this drawback because it essentially forms a recursive density estimation of a data point. It examines whether or not a data sample is located at a strategic location in the input space. The DS method is expressed:
$$V_{M+1} \geq \max_{i=1,...,M} (V_i) \quad (19)$$
where $V_{M+1}$ is the volume of hypothetical rule and $V_i$ is the volume of *i-th* rule. This expression is obtained from $\int_X R_i p(x)dx$ where $p(x)$ is a probability density function assuming that data follow the uniform distribution. Although it encompasses overall contribution of a rule, its uniform data distribution assumption obscures a sample's location.

The DQ method is put forward as another rule growing mechanism to measure fuzzy rule relevance through its density. The difference is, however, observed in the facts that the DQ method involves a weighting factor $U_N$ to address an outlier bottleneck, uses the inverse multi-quadratic function and is well-suited to the multivariable Gaussian function. It adopts similar concept of the recursive density estimation (Angelov et al, .2004) where a rule density is defined as an accumulated distance of a cluster's prototype to all other samples seen so far:
$$DQ_N = \sqrt{\frac{U_N}{U_N(1+a_N)-2b_N+c_N}}$$
$U_N = U_{N-1} + DQ_{N-1}, a_N = X_N \Sigma_N^{-1} X_N, b_N = DQ_{N-1} X_N \alpha_N, \alpha_N = \alpha_{N-1} + \Sigma_N^{-1} X_{N-1}^T, c_N = c_{N-1} + DQ_{N-1} X_{N-1} \Sigma_N^{-1} X_{N-1}^T$
where $\Sigma_N^{-1}$ is an inverse covariance matrix of the hypothetical rule created by a current data point (22)-(24). The DQ rule growing condition is written:
$$DQ_N \geq \max_{i=1,...,M}(DQ_i) \text{ or } DQ_N \leq \min_{i=1,...,M}(DQ_i) \quad (20)$$
Note that $X_N$ in (19) is replaced by the focal point of *i-th* cluster $C_i$ for $DQ_i$. The first condition in (20) reveals a case where a new data point brings the most relevant concept to the training process because it occupies the most populated input region. The second condition in (20) signifies a possible concept change because a data point is situated at a remote region which is outside influence zones of a current rule base. Such samples are vital to introduce new concepts in the current rule base. A new rule is added to the current rule base if both (19) and (20) are satisfied. The growing strategy is consistent with the problematizing part of the Scaffolding theory.

- *Parameterization of a New rule*: once a new rule is added, its parameters have to be initialized. This step warrants in-depth investigation because of the class overlapping case which leads to performance deterioration (Subramanian et al., 2014(a)). That is, a local region is populated by different-classes samples, or a class is not clearly separated. There exist two types of overlapping: inter-class and intra-class. The inter-class overlapping leads to classifier's confusion and imposes a dramatic increase of nonlinearity of decision boundary.

rClass is endued with the class overlapping strategy to ensure a decent input partition of a new rule. At first, the compatibility degree of a new rule with respect to other rules is checked where it can be examined through a firing strength of the winning rule $R_{win}$. The winning rule is here determined by the one with the closest distance to a new rule. Given that it beats a distance threshold $R_{win} \geq threshold$ where *threshold* is chosen as a critical value of $\chi^2$ distribution with $u$ degree of freedom and significance level $\alpha$, the potential per-class concept is executed - $threshold = \exp(-\chi^2(\alpha))$. The potential per-class concept aims to investigate the relationship of a new rule to target classes whether it is more adjacent to the same class than other classes or not. The potential per-class method is mathematically written as follows:

$$\varsigma_o = \sqrt{\frac{(N_o-1)}{(N_o-1)(ab_n+1)+cb_{no}-2bb_{no}}} \quad (21)$$

$ab_n = \sum_{j=1}^{u+m} x_{j,N}^2$, $cb_{no} = cb_{no-1} + \sum_{j=1}^{u+m} x_{j,No}^2$, $bb_{no} = \sum_{j=1}^{m+u} x_{j,N} dd_{j,no}$, $d_{no} = d_{no-1} + x_{No}$.

where $x_{No}$ is the latest sample of the o-th class and $x_N$ is the newest sample, while $N_o$ is the number of samples of the *o-th* class. The class overlapping occurs when the closest class differs from the true class label $\max_{o=1,\dots,C}(\varsigma_o) \neq true\_class\_label$. Based on these two criteria, three scenarios, namely class overlapping, cluster overlapping and non-overlapping, are devised to tackle each possible combination. We start by defining two important variables: $r_{ie}, r_{ia}$ which stand for a distance between a new rule to the closest inter-class rule and the closest intra-class rule respectively. The three conditions are elaborated as follows:

*The class-overlapping condition*: this condition occurs when $\max_{o=1,\dots,C}(\varsigma_o) \neq true\_class\_label$ is met. This implies that a new rule sits in the overlapping region of different-class samples. A new rule is initialised as follows:

$c_{j,M+1} = x_{j,N} - 0.1(c_{j,ie} - x_{j,N}), dist = fac(c_{j,M+1} - c_{j,ie}) \quad (22)$

where *fac* is an overlapping factor determined as a distance ratio between the intra-class cluster and the inter-class cluster $fac = \frac{r_{j,ia}}{r_{j,ie}}$. This setting aims to shrink the coverage of a new rule proportionally to its overlapping degree. Furthermore, the centre of a new rule is shifted away from the inter-class cluster to alleviate the effect of class overlapping.

*The rule-overlapping condition*: this condition portrays a case where a new rule is neighboring to the intra-class cluster. It is pinpointed by $\max_{o=1,\dots,C}(\varsigma_o) = true\_class\_label$. This incurs the risk of overlapping in the rule level which is less harmful than the class-overlapping condition. The following is undertaken to address this condition.

$c_{j,M+1} = x_{j,N} - 0.1(c_{j,ia} - c_{j,ie}), dist = fac(c_{j,M+1} - c_{j,ie}) \quad (23)$

This setting is required to move a new rule away from the intra-class cluster since two rules may move together and end up in the significantly overlapping position. The rule merging strategy is incorporated in the rClass and functions also to handle the rule overlapping case.

*The non-overlapping condition*: this condition delineates a case when a new rule is sufficiently distant from other rules and is indicated by $R_{win} < threshold$. It is safe to construct a new rule:

$$c_{j,M+1} = x_{j,N}, dist = fac(x_{j,N} - c_{j,ia}) \quad (24)$$

A new rule is out of scope of existing rules. A new inverse covariance matrix for all conditions is initialized as $\Sigma_{M+1}^{-1} = (dist^T dist)^{-1}$. A new rule is initialized as a classical cluster in the main axes but rotates after receiving the premise adaptation (28).

The output weight, recurrent weight, and output covariance matrix are crafted as follows:

$$W_{M+1} = W_{win}, \Psi_{M+1} = \omega I, \gamma_{M+1} = \frac{\sum_{i=1}^{M} \gamma_i}{M} \quad (25)$$

where $\omega = 10^5$ is a large positive constant and it assures a convergence toward a batch learning process which has been mathematically proven in (Lughofer, 2010). Moreover, a new weight vector is set as that of the winning rule since it portrays the most compatible concept to the new rule. The recurrent weight is assigned as an average of existing ones to be proportional to the current weights. The self-construction of fuzzy rules here depicts the problematizing facet of the Scaffolding theory because it is capable of addressing rapidly changing process environments by introducing a new rule when needed.

There exist situations where the rule growing criteria in (19) and (20) are not met. Data streams in these conditions merely induce minor conflict to the current belief and do not suffice to trigger the rule adding a scenario. This condition is formalized:

$$V_{M+1} \geq \max_{i=1,\dots,M}(V_i)$$
$$\min_{i=1,\dots,M}(DQ_i) < DQ_N < \max_{i=1,\dots,M}(DQ)_i \quad (26)$$

This condition exhibits a data point having a substantial statistical contribution but is still within the coverage of existing rules. Only the antecedent of the winning rule is refined in this scenario to be well-suited to the currently seen concept. The online adaption of the rule premise is derived from the modified version of sequential maximum likelihood estimation for the multivariate Gaussian function:

$$C_{win} = \frac{N_{win}C_{win}}{N_{win}+1} + \frac{(X-C_{win})}{N_{win}+1} \quad (27)$$

$$\Sigma_{win}^{-1} = \frac{\Sigma_{win}^{-1}}{1-\alpha} + \frac{\alpha}{\alpha+1} \frac{\Sigma_{win}^{-1}((X-C_{win})\Sigma_{win}^{-1}(X-C_{win})^T)}{1+\alpha(X-C_{win})\Sigma_{win}^{-1}(X-C_{win})^T} \quad (28)$$

$$N_{win} = N_{win} + 1 \quad (29)$$

where $\alpha = \frac{1}{N_{win}+1}$. (28) features a direct update of the covariance matrix with the absence of reinversion step and is inspired by the Neuman series (Lughofer et al., 2015). This strategy is desirable since the reinversion step tends to be unstable in the case of an ill-defined matrix and is intractable when dealing with a high input dimension. The adaptation of rule antecedent complements the rule growing scenario to adapt to variations of data streams and functions as the problematizing component of the Scaffolding theory. Only the winning rule is subject to the tuning phase to prevent the overlapping situation and to the fact that it is the closest one to the current sample. The movement of the winning rule decreases as the increase of its supports since a cluster is encouraged to capture more supports initially but should be made less sensitive at the later stage to avoid sample redundancies. Note that the winning rule can be determined as the closest rule or that of the highest posterior probability (Pratama et al, 2015(a)).

- *Rule Pruning Strategy*: rClass is equipped by two rule pruning scenarios, namely the Extended Rule Significance (ERS) method and the Potential+ (P+) method, which aim to simplify the rule base complexity as well as to prevent the overfitting. The ERS method shares the same principles as the DQ method, but it quantifies the statistical contribution of existing rules instead. It targets superfluous rules which play little during their lifespan. The P+ method is designed to detect obsolete rule which is no longer relevant to the current training concept. It also functions as a rule recall scenario, which overcomes the recurring drift. The DQ method is mathematically expressed as follows:

$$ERS_i = \sum_{o=1}^{C} \sum_{j=1}^{2u+1} w_{i,o} \frac{V_i}{\sum_{i=1}^{M} V_i} \quad (30)$$

It is also seen from (30) that the ERS method also takes into account the contribution of the output weight. A fuzzy rule is pruned provided that $ERS_i < \widehat{ERS_i} - ERS_{i,\sigma}$ condition is satisfied where $\widehat{ERS_i}, ERS_{i,\sigma}$ are mean and standard deviation of the ERS of the *i-th* rule. The ERS method realises the fading of the Scaffolding theory which aims to alleviate the rule base complexity. The P+ method is formulated as follows:

$$\aleph_{i,n} = \sqrt{\frac{(N-1)\aleph_{i,n-1}^2}{(N-1)\aleph_{i,n-1}^2+(N-2)\left(1-\aleph_{i,n-1}^2\right)+\aleph_{i,n-1}^2 d_{i,n}}} \quad (31)$$

where $d_{i,n}$ is the Mahalabobis distance between a newly observed sample and the focal point of *i-th* rule. $\aleph_{i,0}$ is initialized as zero. As with the ERS method, a fuzzy rule is pruned if this condition $\aleph_{i,n} < \widehat{\aleph}_{i,n} - \aleph_{i,n}^\sigma$ is met. This condition aims to check decline of the P+ value of *i-th* rule during its lifespan and portrays a situation when a rule loses its relevance. The P+ method is in line with the problematizing of the Scaffolding theory because it deals with obsolete rules due to concept change.

- *Rule Recall Strategy*: the recurring or cyclic drift pinpoints a condition where an old concept reappears again in the future. This issue causes a previously pruned rule to be valid again in the future (Pratama et al., 2015(a)) and such rules should be recalled to cope with current data distribution instead of creating a new rule from scratch because it may impose catastrophic forgetting of previously learned concept. The P+ method also functions to reactivate previously pruned rule because it can precisely monitor the relevance of fuzzy rules by inspecting the evolution of local density. An old rule should be recalled when its relevance beats any existing rule including that of the hypothetical rule as follows:

$$\max_{i*=1,\ldots,P*}(\aleph_{i*,n}) > \max_{i=1,\ldots,P+1}(DQ_{i,n}) \quad (32)$$

where *P\** is the number of rules discarded by the P+ method. It is observed from (23) that the P+ method can be compared directly with the DQ method because both of which estimates the local density. The P+ method is, however, updated when a new sample is learned. The rule recall mechanism is undertaken as follows:

$$C_{P+1} = C_{i*}, \Sigma_{P+1}^{-1} = \Sigma_{i*}^{-1}, W_{P+1} = W_{i*}, \Psi_{P+1} = \Psi_{i*} \quad (33)$$

Although previously pruned rules are retained in the memory, computational complexity and memory demand still diminish since old rules are excluded from other learning scenarios.

- *Cluster Splitting Mechanism*: an over-sized cluster undermines the generalization capability of a model because it opens a chance for one cluster to cover two or more distinct data clouds. In the presence of local drift, it causes a cluster to blow-up (over-sized) because of gradual change of distribution in one local region. This situation calls for the rule splitting mechanism which divides a cluster to two disjoint cluster subject to the following condition.

$$V_{win} > \delta \sum_{i=1}^{M} V_i \quad (34)$$

where $\delta \in [0.5, 0.9]$ is a tolerance threshold which steers the intensity of splitting process and is fixed at 0.8 in all our simulations. Furthermore, only the winning rule is to be checked in (34) because it is the only cluster receiving the adaptation. The splitting mechanism is defined as follows:

$$C_{M+1} = C_{win} \pm (g_{max}\sqrt{\alpha_{max}} + q\sqrt{\alpha_{max}})$$
$$\Sigma_{M+1}^{-1} = \Sigma_{win}^{-1} - ((g_{max}\sqrt{\alpha_{max}})^T(g_{max}\sqrt{\alpha_{max}}))^{-1}$$
$$W_{M+1} = W_{win}, \Psi_{M+1} = \Psi_{win}, N_{M+1} = \frac{N_{win}}{2}$$

where $\alpha_{max}, g_{max}$ are the largest eigenvalue and its corresponding eigenvector, while *q* is a predefined constant which controls a distance of two clusters of the splitting mechanism. Because the eigenvalue signifies the variance of $\Sigma_{win}$ toward the direction of eigenvector, the maximum eigenvalue is selected to prevent the loss of orientation and population since it indicates the underlying direction of a cluster. The rule splitting mechanism is consistent with the problematizing of the Scaffolding theory.

- *Rule Forgetting Mechanism*: the rule forgetting mechanism is an effective avenue in handling the gradual concept drift because it strengthens the evolution of a cluster to follow the concept drift. It is also applicable to cope with the local concept drift because unique forgetting levels for every local region are assigned in accordance with drift intensity and velocity. Forgetting levels for every local region are determined:

$$\lambda_i = 1 - 0.1(\aleph_{i,n} - \aleph_{i,n-1}) \quad (35)$$
$$\lambda_{trans,i} = -9.9\lambda_i + 9.9 \quad (36)$$
$$N_i = N_i - N_i \min(\lambda_{trans,i}, 0.99) \quad (37)$$

The forgetting level $\lambda_i \in [0.9,1]$ is measured from the rate of the P+ values (31) in two consecutive measurements where the lower the value indicates a stronger forgetting level. Since the P+ method is concerned with the local density, a change of local density discloses an indication of a local concept change. The local concept drift is handled in both input and output space where the forgetting level is inserted in the FWGRLS formulate to deal with the local concept change in the output space while a reduction of cluster population (28) is performed to overcome the concept drift in the input space.

- *Rule Merging Strategy*: The rule merging scenario is realized using the Bhattacharyya distance and the blow-up check (Lughofer, 2015(a)). The advantage of the Bhattacharyya similarity measure lies in its aptitude to approximate the spread of the multivariate Gaussian distribution which is equivalent to the multivariable Gaussian function. It is furthermore threshold-free which is appealing in dealing with online data streams. The blow-up check copes with the so-called cluster delamination which delineates an over-sized cluster covering two or more distinguishable data clouds. Such cluster hinders the model's generalization because the specificity of the cluster lowers. The Bhattacharyya similarity measure is expressed as follows:

$$S_{win,i} = \frac{1}{8}(C_{win} - C_i)\Sigma_{comb}^{-1}(C_{win} - C_i)^T$$
$$+ \frac{1}{2}\ln(\frac{\det(\Sigma_{comb}^{-1})}{\det(\Sigma_{win}^{-1})\det(\Sigma_i^{-1})})$$

where $\Sigma_{comb}^{-1} = \frac{\Sigma_{win}^{-1} + \Sigma_i^{-1}}{2}$. $S_{win,i} = 0$ indicates that two clusters are touching and two clusters are overlapping when it results in a positive value $S_{win,i} > 0$ while a negative value $S_{win,i} < 0$ signifies that two clusters are disjoint.

The blow-up check is carried out by inspecting the volume of a merged cluster whether it beats a total volume of two independent clusters. This aims to assess whether two clusters are homogeneous because two non-homogenous clusters lead to an over-sized merged cluster which is prone to the cluster delamination condition. We arrive at the following condition in coalescing two clusters:

$$S_{win,i} > 0, V_{merged} \leq u(V_{win} + V_i) \quad (29)$$

$u$ is inserted in (29) to deal with the curse of dimensionality. The rule merging mechanism is undertaken provided that (29) is satisfied. The merging process is performed using the weighted average strategy:

$$C_{merged} = \frac{C_{win}N_{win} + C_i N_i}{C_{win} + C_i} \quad (30)$$
$$\Sigma_{merged}^{-1} = \frac{\Sigma_{win}^{-1}N_{win} + \Sigma_i^{-1}N_i}{N_{win} + N_i} \quad (31)$$
$$N_{merged} = N_i + N_{win} \quad (32)$$

The weighted average strategy allows a rule with a higher support to be more influential to the final shape of a merging result. The rule consequent is merged by taking into account the degree of contradiction between the rule antecedent and the rule consequent because a contradiction exists when the two rule consequents are dissimilar, but their antecedents are similar. The similarity of rule consequents is first measured where it is pinpointed by the angle created between two rule consequents in the output space. Note that we just consider the linear terms of the rule consequent while excluding their higher order terms as follows:

$$S_{win,i}^{out} = \begin{cases} 1 - \frac{2}{\pi}\phi, \phi \in [0, \frac{\pi}{2}] \\ \frac{2}{\pi}(\phi - \frac{\pi}{2}), \phi \in [\frac{\pi}{2}, \pi] \end{cases} \quad (33)$$

$$\phi_o = \arccos(\frac{a_o^T b_o}{|a_o||b_o|}) \quad (34)$$

$$\phi = \max_{o=1,\dots,C}(\phi_o) \quad (35)$$

$a_o = [w_{0,win}^o, w_{1,win}^o, w_{3,win}^o, \dots, w_{u,win}^o]$
$b_o = [w_{0,i}^o, w_{1,i}^o, w_{3,i}^o, \dots, w_{u,i}^o]$

It is worth noting the higher order term of the Chebyshev polynomial merely fashions its nonlinear oscillation while its orientation is steered by its linear terms. This similarity measure is referred to when merging the rule consequent. The old rule consequent is retained when the similarity of the rule consequent is lower than the similarity of the rule consequent. The merging process is carried out using the Yager's participatory learning inspired approach as follows:

$$\Omega_{merged} = \Omega_1 + \gamma\delta(\Omega_1 - \Omega_2) \quad (36)$$
$$\gamma = \frac{N_1}{N_1 + N_2} \quad (37)$$
$$\delta = \begin{cases} 1, S_{win,i}^{out} \geq S_{win,i} \\ 0, S_{win,i} > S_{win,i}^{out} \end{cases} \quad (38)$$

About the Yager's participatory learning scheme, $\gamma \in [0,1]$, $\delta$ can be regarded as the basic learning rate and the compatibility index between two models respectively, while the arousal index is set constant at 0. $\Omega_1$ here denotes the rule consequent of the dominant rule having more supports than $\Omega_2$, $N_1 > N_2$. The rule merging process focusses on the winning rule because it has a higher risk of overlapping than others due to the rule antecedent learning (27) - (29). The rule merging strategy represents the complexity reduction of the Scaffolding theory.

- *Input Weighting Strategy*: The online feature weighting scenario based on the L1-norm Fisher Discriminant Analysis (FDA) is integrated into rClass. This differs from an earlier version of the online feature weighting utilizing the L2-norm (Lughofer, 2011) leading to at least three problems: outliers, rank limit, and small sample size. The L1 norm FDA was proposed in (Wang et al., 2014) and its cost function is written as follows:

$$J(\omega) = \frac{\sum_{o=1}^{C} N_o |\omega(\bar{x}_o - \bar{x})^T|}{\sum_{o=1}^{C} \sum_{n=1}^{N_o} |\omega(x_{n,o} - \bar{x})^T|} \quad (39)$$

where $\omega \in \Re^{1 \times u}$ is a transformation vector. The online version of the L1 norm FDA is derived by formulating recursive expressions of the mean of the *o-th* class data samples $\bar{x}_o$, the mean of the data samples $\bar{x}$ as follows:

$$\bar{x}_o = \frac{(N_o - 1)}{N_o} \bar{x}_o + \frac{x_{No.o}}{N_o}, \bar{x} = \frac{(N-1)}{N} \bar{x} + \frac{x_N}{N} \quad (40)$$

With $x_N$ the latest (the *N*-th) data sample loaded and $x_{No.o}$ the latest (the *No*-th) data sample from class *o*. The online version of within-class scatter matrix is formalized as follows:

$$\sum_{o=1}^{C} \sum_{n=1}^{N_o} |\omega(x_o - \bar{x})^T| = \sum_{o=1}^{C} \omega \Sigma_o \quad (41)$$
$$\Sigma_o = \Sigma_o + |x_o - \bar{x}| \quad (42)$$

The projection vector $\omega$ is fine-tuned using the gradient ascent procedure maximizing the cost function (39) as follows:

$$\omega = \omega + \eta \frac{\partial J(\omega)}{\partial \omega} \quad (43)$$

where $\eta = 10^{\wedge}(-3)$ is a learning rate. The first order derivative of the cost function with respect to the projection vector is defined as follows:

$$\frac{\partial J(\omega)}{\partial \omega} = \frac{\sum_{o=1}^{C} s_o N_o (\bar{x}_o - \bar{x})}{\sum_{o=1}^{C} N_o |\omega(\bar{x}_o - \bar{x})|} - \frac{\sum_{o=1}^{C} r_o \Sigma_o}{\sum_{o=1}^{C} \omega \Sigma_o} \quad (44)$$

The denominator sometimes goes to zero quickly. This is addressed by replacing it with $\omega = \omega + \Delta\omega / \|\omega + \Delta\omega\|$ where $\Delta\omega$ is a small nonzero random vector if the denominator becomes too small. $s_o, r_o$ are two sign functions:

$$s_o = sgn(\omega(\bar{x}_o - \bar{x})), r_o = sgn(\omega(x_o - \bar{x}_o)) \quad (45)$$

The input weight is produced by the Leave-One-Feature-Out (LOFO) scheme which studies the contribution of an input feature by masking it for the training process. An important input attribute should have a high discriminatory power. Therefore, when it is ruled out, the cost function (39) should significantly reduce. The input weight is then quantified as follows:

$$\lambda_j = 1 - \frac{J_j - \min_{j=1,\ldots,u}(J_j)}{\max_{j=1,\ldots,u}(J_j) - \min_{j=1,\ldots,u}(J_j)} \quad (46)$$

where $J_j$ is the value of the cost function (39) when an input attribute *j* is ignored from the training process. The input weight $\lambda_j$ is involved in all parts of learning process: inference scheme, etc. The online weighting mechanism is in line with the complexity reduction of the Scaffolding.

- *Parameter Learning Strategy*: rClass utilizes the so-called Fuzzily Weighted Generalized Least Square (FWGRLS) method to fine-tune the rule consequent. The FWGRLS forms a local learning version of the GRLS method (Xu et al., 2006) which puts forward a weight decay term in the cost function of the RLS method. The weight decay term navigates the weight vector such that it hovers around in a small and bounded interval. Such property is vital for the generalization's capability of a model and compactness of a rule base since a superfluous rule usually tends to have a very small rule consequent. This issue eases the task of the ERS method (22) in capturing inconsequential rules. The FWGRLS method is formulated:

$$K(N) = \Psi_i(N-1)F(n)(\frac{\lambda_i \Delta(N)}{\Lambda_i(N)} + F(N)\Psi_i(n-1)F(N)^T)$$
$$\Psi_i(N) = \Psi_i(N-1) - K(N)F(n)\Psi_i(N-1)$$
$$W_i(N) = W_i(N-1) - \omega \Psi_i(N) \nabla \xi(W_i(N-1)) + \Psi_i(N)(t(N) - y(N))$$
$$y(N) = x_e W_i, F(N) = \frac{\partial y(N)}{\partial \Psi_i(N)} = x_{en}$$

where $\Delta(N)$ is a Hessian matrix whose diagonal elements consist of the spatial firing strength $R_i$ and $\Lambda_i(N)$ is a covariance matrix of the modelling error simply set as the Hessian matrix. $\lambda_i$ is the forgetting factor determined through the rate of change of the P+ values in two consecutive instants (31). Note that the forgetting factor aims to intensify the movement of the rule consequent in the presence of concept drift. $K(N)$ is the Kalman gain and $\Psi_i$ is the output covariance matrix which is unique per rule because of the local learning principle. The adaptation process is carried out separately per rule as the local learning principle. This strategy provides flexibility because the training process of a rule does not affect the stability and convergence of other rules. $\nabla \xi(W_i(N-1))$ is the gradient of the weight decay function where the quadratic weight decay function is selected because it is capable of shrinking the weight vector proportionally from its current values. $\omega$ is a predefined constant which governs the dominance of the weight decay term and set as its default value $10^{-3}$.

The recurrent weight of rClass is adjusted using the Zero Error Density Maximization (ZEDM) method (Subramanian et al., 2013), which modifies the classical gradient descent method. The ZEDM method replaces the cost function of the gradient descent method with the error entropy concept to attain a reliable approximation of high order statistical behavior. The goal of the ZEDM method is to minimize the distance between the probability distribution of the target class and the classifier's output. This strategy is equivalent to forcing the system's error toward zero. Because the exact expression of the error entropy is difficult to be modeled, the Parzen Window estimation is put forward as follows:

$$f(0) = \frac{1}{Nh\sqrt{2\pi}} \sum_{n=1}^{N} \exp(-\frac{e_{n,h}^2}{2h^2}) \quad (47)$$

where $N$ denotes the number of training samples seen thus far, and $h$ stands for a smoothing parameter set as one while $e_{n,o}$ is the system error at the $n$-th observation and $o$-th class label. The optimization procedure is carried out using the gradient descent method as follows:

$$\gamma_{i,o}(N) = \gamma_{i,o}(N-1) - \eta \frac{1}{N\sqrt{2h}} \sum_{n=1}^{N} \exp\left(-\frac{e_{n,h}^2}{2}\right) \frac{\partial E}{\partial \gamma_{i,o}} \quad (48)$$

where $\eta$ denotes the learning rate. (30), however, revisits preceding training samples which is intractable for the life-long learning case. It is therefore modified as follows:

$$\sum_{n=1}^{N} \exp\left(-\frac{e_{n,h}^2}{2}\right) = A_N = A_{N-1} + \exp\left(-\frac{e_{n,h}^2}{2}\right)$$

The gradient term in (30) is derived using the chain rule as follows:

$$\frac{\partial E}{\partial \gamma_{i,o}} = \frac{\partial E}{\partial y} \frac{\partial y}{\partial \varphi_{i,o}} \frac{\partial \varphi_{i,o}}{\partial \gamma_{i,o}}$$

$$\frac{\partial E}{\partial \gamma_{i,o}} = (y_o - t_o)(R_i - \varphi_{i,o}) \frac{(x_e W_{i,o} - y_0)}{\sum_{i=1}^{M} \varphi_{i,o}}$$

The learning rate is well-known to greatly influence the convergence of learning process. A stable interval of the learning rate is derived using the Lyapunov stability criterion as follows:

$$0 < \eta < \frac{2N\sqrt{\pi}}{(P_{o,max})^2 A_N} \quad (49)$$

where $P_{o,max} \approx \frac{1}{M}$. This interval, however, does not guarantee fast convergence because of its static nature. This drawback, usually, warrants an adaptive learning rate as follows:

$$\eta = \begin{cases} \rho_1 \eta, f(0)_n \geq f(0)_{n-1} \\ \rho_2 \eta, f(0)_n < f(0)_{n-1} \end{cases} \quad (50)$$

where $\rho_1 \in (1,1.5], \rho_2 \in [0.5,1)$ are learning rate factors which steer the fluctuation of learning rates. The learning rate goes up when the cost function augments to expedite the convergence rate. The learning rate decreases when the cost function lessens to mitigate the convergence rate. The parameter learning scenario actualises the problematizing of the Scaffolding theory as it contributes in handling the concept drift.

*B.3 When-to-Learn*

The when-to-learn strategy encompasses a sample coming through the what-to-learn strategy but does not satisfy the sample learning criteria in (19), (20), and (26). Such sample is set as reserved samples and intended for future use because it may be fruitful to refine the rule base for possible uncovered states of already seen training samples. These samples are learned when all training samples have been finished or in the realm of data streams when the system is idle, and no new samples are received.

**4. Problem Descriptions**

In this section, the experimental setup and the design of the evolving intelligent system is discussed. The experimental setup is the same as described in (Dimla et al., 2000) but is detailed here for the convenience of readers. The machine used in the experiment was a variable speed centered lathe of type Lang Swing J6.

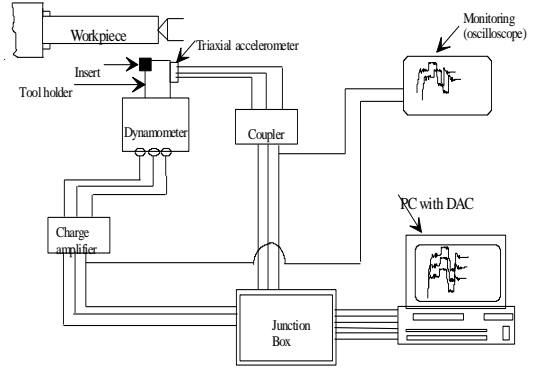

Figure 4. The Schematic of experimental test rig (Dimla et al, 2000)

The two tool inserts used, P15 and P25, were cemented carbide coated via chemical vapor deposition and consisted of grades of 'throw-away' indexable inserts with integral chip-breaker geometry, held in place by a negative rake tool holder. P15 had a thick wear resistant coating on a hard, resistant substrate while P25 had a thick ($\approx 10\mu m$) layer of $Al_2O_3$ on top of a medium size titanium carbon nitride (TiCN) giving it a high wear resistance. No cutting fluid was used. The workpiece material used was low carbon alloy steel of the EN24 type (oil quenched, rolled and tempered) which is relatively hard to accelerate tool-wear at the expense of a shorter tool life.

The sensors for the vibrations and cutting forces were a tri-axial Kistler mini accelerometers (type 8730A) for acceleration signal measurement and a Kistler tool post dynamometer platform (type 9263A) for cutting force measurement in three planes respectively. The signals from the sensors were passed through a signal conditioning unit and various peripheral signal conditioning instruments as shown in the schematic diagram (Figure 4). Sampling was performed at a frequency of 30 kHz while recording 4096 data samples per channel.

In the traditional TCM, interrupted test cuts were conducted at fixed cutting conditions with fresh tool inserts until failure or when wear levels had accumulated at which continued cutting risked catastrophic failure. The metacognitive TCM offers a flexible avenue where the training process was carried out on the fly and importantly incurs low labelling cost because each cutting process does not necessarily lead to manual labelling effort by operator through visual inspection of cutting condition. Due to the rate of tool wear, cuts normally lasted less than 10 seconds at the beginning of each run. As the cutting process progressed, the duration of each cut was systematically increased to well over 30 seconds for complete stabilization of the cutting process to be achieved while at the same time allowing significant tool wear accumulation.

We assigned different machining parameters in terms of cutting speed, feed-rate and depth of cut for each cut to suit the production requirement. During each run, the cutting force and vibration picked up by the force sensor and accelerometer were recorded online and generated data streams. When operator's intervention is needed to feed the ground truth of the cutter's condition, the cutting tool is manually inspected – with the flank and nose being measured. The measurement is then compared against a pre-set threshold to determine the state of the tool, as follows (Dimlan et al., 2000):

Table 1: Examples of machining parameters, measured signals, and tool wear

| Cutting speed | Feed-rate | Depth of cut | Static Force X | Static Force Y | Static Force Z | Dynamic Force X | Dynamic Force Y | Dynamic Force Z | Acceleration X | Acceleration Y | Acceleration Z | Flank | Nose | Chipped |
|---|---|---|---|---|---|---|---|---|---|---|---|---|---|---|
| 0.857 | 0.5 | 0.571 | 0.604 | 0.603 | 0.726 | 0.475 | 0.296 | 0.43 | 0.545 | 0.502 | 0.485 | 1 | 1 | 1 |
| 0.786 | 0.75 | 0.571 | 0.133 | 0.088 | 0.682 | 0.681 | 0.394 | 0.632 | 0.748 | 0.537 | 0.546 | 0 | 0 | 0 |
| 0.786 | 0.25 | 0.571 | 0.087 | 0.063 | 0.261 | 0.235 | 0.097 | 0.169 | 0.217 | 0.147 | 0.118 | 1 | 0 | 0 |
| 0.786 | 0.25 | 0.571 | 0.086 | 0.061 | 0.276 | 0.167 | 0.083 | 0.128 | 0.191 | 0.133 | 0.113 | 0 | 0 | 0 |

- 000 – nominally sharp
- 100 – high flank wear
- 010 – high nose wear
- 001 – chipped/fractured nose
- 110 – high flank and high nose wear
- 111 – high flank and chip / fractured nose

The aforementioned pre-set threshold is as follows:
- Flank wear mark value ≤ 0.15mm, tool insert nominally sharp
- Flank wear mark value > 0.15mm, tool insert worn (high flank)
- Nose wear length ≤ 0.2mm, nominally sharp
- Nose wear length > 0.2mm, tool worn (nose fractured / chipped)

Table 2. Consolidated Numerical Results of Benchmarked Algorithms

| Algorithms | CR | RT | NS | FR |
|---|---|---|---|---|
| rClass | 0.84 | 0.56 | 26 | 2 |
| eClass | 0.75 | 1.3 | 50 | 10 |
| pClass | 0.75 | 0.65 | 50 | 1 |
| GENEFIS-class | 0.78 | 0.6 | 50 | 10 |
| gClass | 0.71 | 0.68 | 50 | 2 |
| IT2McFIS | 0.83 | 0.7 | 36 | 9 |

CR: Classification Rate, RT: Runtime, NS: Number of Samples, Fuzzy Rules

The monitoring process was carried out fully in the online mode with an intermittent stoppage to label data streams. Note that the cutting force and vibration are continuously recorded during each run. The advantage of rClass is seen in the absence of a retraining phase when a new data record is observed. All training samples are processed on the fly without revisiting preceding samples. Examples of historical data captured in the experiment is shown in Table 1. Note: The depth of cut is an important parameter for the experiment because a deeper cut will naturally induce higher cutting force. To avoid confusion as to whether the increase in cutting force is due to worn tools or depth of cut, the latter should be included as the input parameter. Besides the measured cutting force and vibration, other signals such as spindle current, acoustic emission and temperature can be added as inputs to the predictive analytics in the future.

Table 3. Consolidated Numerical Results of Benchmarked Algorithms

| Algorithms | CR | RT | NS | FR |
|---|---|---|---|---|
| rClass | 0.84 | 0.05 | 26 | 2 |
| eClass | 0.71 | 0.28 | 50 | 11 |
| pClass | 0.77 | 0.35 | 50 | 1 |
| GENEFIS-class | 0.83 | 0.45 | 50 | 8 |
| gClass | 0.8 | 0.74 | 50 | 2 |
| IT2McFIS | 0.88 | 0.34 | 34 | 14 |

CR: Classification Rate, RT: Runtime, NS: Number of Samples, Fuzzy Rules

**5. Experiment**

The input of the EIS are the first 12 columns of Table 1: cutting speed, feed-rate, depth of cut, static force (X,Y,Z), acceleration (X,Y,Z), while the remaining three columns are the cause of tool wear observed during the experiment. It is worth noting that there are three factors of the tool wear, namely flank wear, nose wear and chipped or fractured nose. The tool wear is attributed by a combination of the three factors: high flank wear and chip/fractured nose. Our dataset forms a four-classed multiclassification problem where the target classes covers four conditions: nominally sharp, high flank wear, high flank and nose wear, high flank wear and chipped/fractured nose. 50 samples were used to build our hypothesis, and subsequent 69 samples were utilized as the testing samples. rClass was compared against five state-of-the-art classifiers:

- eClass (Angelov et al., 2008(a)) is a prominent evolving classifier in the literature, which forms an extension of eTS (Angelov, 2004) for classification problems. It takes advantage of eClustering to cluster the input space on the fly.

- GENEFIS-class (Pratama et al., 2014(c)) is a modified version of GENEFIS to deal with the classification problem. It actualises a holistic concept of the EIS where a rule can be generated, merged, pruned on demands, while it is equipped with an online feature selection.
- pClass (Pratama et al., 2015(a)) is an evolving classifier which enhances the GENEFIS-class. It targets the uniform distribution assumption of the GENEFIS-class and is capable of addressing the recurring drift due to its rule recall scenario.
- gClass (Pratama et al., 2015(b)) is a metacognitive classifier, which works fully in the single-pass learning mode. It goes beyond the standard metacognitive learning by introducing the Scaffolding theory into the how-to-learn phase. It can be seen as a semi-supervised learner as a result of the online active learning scenario.
- McITSFIS (Subramanian, 2014(a)) is a metacognitive classifier which combines the theory of interval type-2 fuzzy system into the metacognitive learning. It modifies McFIS (Subramanian, 2014(b)) into the interval type-2 fuzzy system.

All benchmarked algorithms are structured in the MIMO architecture. The consolidated algorithms are evaluated against four evaluation criteria: classification rate, the number of hidden nodes, the number of network parameters and the number of samples. Numerical results of benchmarked algorithms are summed up in Table 2.

The advantage of rClass over its counterpart is obvious in Table 2 where it delivered the most encouraging accuracy while attaining the lowest complexity in terms of both the number of rule and the execution time. rClass achieved this result by exploiting the lowest number of samples confirming the success of the active learning scenario.

We also simulated the performance of consolidated algorithms under a reduced input dimension – only 7 input attributes. Cutting speed, deep of cut, which happen to be fixed machining parameters, are excluded. Figure 6(a) displays the trace of fuzzy rule during the training process, while Figure 6(b) visualizes the evolution of feature weights.

Under a reduced input dimension, rClass still produced consistent performance with negligible deterioration in accuracy. Significant performance improvement is seen in IT2McFIS but it should be noted that it comes with higher complexities – number of rule, execution time and training samples. The self-evolving property of the rClass is confirmed in Figure 6(a) where it starts its learning process from scratch with an empty rule base and fuzzy rules are grown, pruned and merged flexibly during the training process. Figure 6(b) demonstrates the feature weighting capability of rClass where feature weights can be dynamically assigned to input attributes in accordance with their significance to the training process. The feature weighting strategy induces the so-called soft dimensionality reduction minimizing the impact of poor input attributes. It differs from the hard input selection paradigm which causes discontinuity of the training process because once pruned an input attribute cannot be retrieved. The feature weighting mechanism also contributes positively to the compactness of the rule base since it results in a small distance to inconsequential input attributes.

To further investigate the consistency of consolidated algorithms, another series of experiments using historical data instead of data streams was undertaken. Here 50-fold random permutation was followed where data are randomly shuffled to avoid data-order dependency. The data proportion for training and testing remained the same as the online scenario. Table 4 tabulates consolidated numerical results.

From Table 4, it is seen that rClass delivered comparable predictive accuracy with significantly less complexity in terms of NS, FR, and RT. It is understood that this problem has a small sample size and this causes no significant difference in accuracy between the metacognitive classifier and the evolving classifier. However, rClass and gClass impose low labeling cost because not all samples are subject to a labeling effort.

Table 4. Consolidated Numerical Results of Benchmarked Algorithms

| Algorithms | CR | RT | NS | FR |
|---|---|---|---|---|
| rClass | 0.94±0.21 | 0.01±0.01 | 24.4 | 2.4±0.5 |
| eClass | 0.92±0.03 | 0.11±0.02 | 50 | 9.4±1.9 |
| pClass | 0.93±0.007 | 0.01±0.005 | 50 | 3 |
| GENEFIS-class | 0.91±0.05 | 0.14±0.03 | 50 | 11.5±8.9 |
| gClass | 0.93±0.15 | 0.33±0.11 | 32.3 | 3.9±1.1 |
| IT2McFIS | 0.93±0.21 | 0.01±0.005 | 34.4 | 13.3±3.6 |

CR: Classification Rate, RT: Runtime, NS: Number of Samples, Fuzzy Rules

As with in the online procedure, we tested benchmarked algorithms with a lower input dimension. Seven input variables, namely feed rate, static components of cutting force in three cutting axis (X,Y,Z), dynamic components of cutting force in two cutting axis (X,Z) and accelerometer signal in Y axis are extracted to perform predictive analytics. Numerical results are tabulated in Table 5.

Table 5. Consolidated Numerical Results of Benchmarked Algorithms

| Algorithms | CR | RT | NS | FR |
|---|---|---|---|---|
| rClass | 0.93±0.2 | 0.05±0.03 | 32.4 | 3.2±2.9 |
| eClass | 0.9±0.04 | 0.22±0.08 | 50 | 9.2±2.3 |
| pClass | 0.88±0.00 | 0.09±0.09 | 50 | 4 |
| GENEFIS-class | 0.91±0.02 | 0.32±0.17 | 50 | 6.9±4.7 |
| gClass | 0.71±0.14 | 0.34±0.16 | 33.5 | 3.8±1.3 |
| McIT2FIS | 0.91±0.11 | 0.38±0.16 | 34.6 | 13.3±3.4 |

CR: Classification Rate, RT: Runtime, NS: Number of Samples, Fuzzy Rules

It is observed that no significant performance difference exists with a lower dimensional dataset. rClass consistently outperforms its counterparts in all evaluation criteria. Notwithstanding that McIT2FIS consumed the same number of samples, it charged expensive labeling cost because the sample selection scenario was undertaken with full manual intervention.

## 6. Conclusion

A metacognitive approach to tool condition monitoring is proposed. It is designed using a recently published metacognitive learning algorithm, namely Recurrent Classifier (rClass). rClass presents a synergy between the metacognitive model and the Scaffolding theory which brings a step closer to a plug-and-play learner. The efficacy of rClass for PHM was investigated in the tool wear prediction of a variable speed centered lathe of type Lang Swing J6. It was compared against state-of-the-art classifiers, and it was found that rClass produced the most encouraging accuracy, while retaining the most compact and parsimonious structure. Furthermore, rClass offered lower labeling cost and fast training speed than its counterparts. Our future work will be devoted toward improving the robustness of our algorithm against noisy data and missing values.

**Acknowledgements**

This work is fully supported by the LTU start-up grant. The fourth author acknowledges the Austrian research funding association (FFG) within the scope of the 'IKT of the future' programme (contract # 849962), as well as the Austrian COMET-K2 programme of the Linz Center of Mechatronics (LCM).

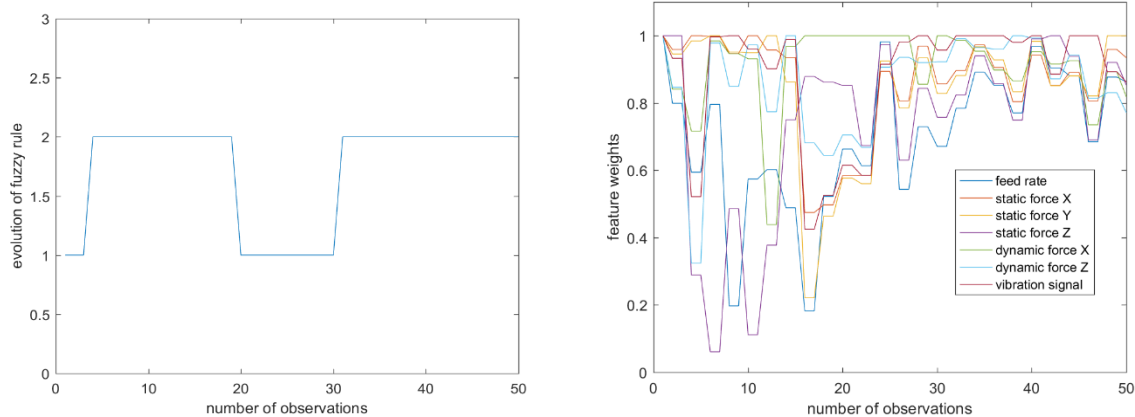

Figure 6: (a) fuzzy rule evolution during the training process, (b) evolution of feature weights during the training process